\documentclass{bmvc2k}
\usepackage{times}
\usepackage{epsfig}
\usepackage{graphicx}
\usepackage{amsmath}
\usepackage{amssymb}
\usepackage{url}
\usepackage{amsfonts}
\usepackage{amssymb}
\usepackage{makeidx}
\usepackage{booktabs}
\usepackage{latexsym}
\usepackage{algorithm}
\usepackage{algpseudocode}
\usepackage{multirow}

\title{Semi-Bagging Based Deep Neural Architecture to Extract Text from High Entropy Images}

\addauthor{Pranay Dugar}{catchpranay@gmail.com}{1}
\addauthor{Anirban Chatterjee}{achatt07@gmail.com}{1}
\addauthor{Rajesh Bhat}{rajeshbhatpesit@gmail.com}{1}
\addauthor{Saswata Sahoo}{ssaswata@yahoo.co.in}{1}

\addinstitution{
 Walmart Labs, \\
 Bangalore, India
}
\runninghead{}{}

\def\etal{\emph{et al}\bmvaOneDot}

\begin{document}

\maketitle

\begin{abstract}
    Extracting texts of various size and shape from images containing multiple objects is an important problem in many contexts, especially, in connection to e-commerce, augmented reality assistance system in natural scene etc. The existing works (based on only CNN) often perform sub-optimally when the image contains regions of high entropy having multiple objects.
    This paper presents an end-to-end text detection strategy combining a segmentation algorithm and an ensemble of multiple text detectors of different types to detect text in every individual image segments independently. The proposed strategy involves a super-pixel based image segmenter which splits an image into multiple regions. A convolutional deep neural architecture is developed which works on each of the segments and detects texts of multiple shapes, sizes and structures. It outperforms the competing methods in terms of coverage in detecting texts in images especially the ones where the text of various types and sizes are compacted in a small region along with various other objects. Furthermore, the proposed text detection method along with a text recognizer outperforms the existing state-of-the-art approaches in extracting text from high entropy images. We validate the results on a dataset consisting of product images on an e-commerce website.
\end{abstract}
%-------------------------------------------------------------------------

\section{Introduction}
Text detection from an image having multiple texts and other objects has been a very important problem in recent times due to the recent advancement and increasing application of computer vision techniques in different domains, for example, e-commerce, automatic driving assistance system, security camera surveillance etc. Detecting regions of texts and extracting that information from natural images is a challenging problem due to the presence of multiple texts of various shapes and sizes and many other objects together. One important problem in this context is extracting texts from product images, particularly in e-commerce, especially, when the intent is to extract brand, product type, various attributes from the product label.  Large e-commerce companies sell billions of products through their websites. All these products are associated with one or more product images containing various textual information about them. Extracting this information are important not only to enhance the quality of the product catalogue but also to examine that product information with respect to various compliance policies of the organization. One of the primary requisites to extract information from product images is to extract text from those images with high accuracy and coverage. 
\par
A vast amount of literature is already available regarding the detection and extraction of text from an image. The problem of text detection in an image has been extensively studied from different perspectives. Neuman \etal~\cite{Neuman2012} proposed a technique to detect individual characters in an image and then group them into words depending on various attributes of the detected characters. Jaderberg \etal~\cite{Jaderberg2016} proposed a technique to directly detect the words in an image. Zhang \etal~\cite{Zhang2015} proposed a method to detect a text line in a scene image directly and then splitting that line into words. TextBoxes~\cite{TB2016} is a word-based method to detect text in a scene image. Unlike Jaderberg \etal~\cite{Jaderberg2016}, which comprises three detection steps and each further includes more than one algorithm, the algorithm of {\em Textbox} is much simpler. It requires only one network to be trained end-to-end. {\em Textbox++}~\cite{TBPP2018} is an improvement over {\em Textbox} and it offers the capability to detect oriented text as well. However, both {\em Textbox++} and {\em Textbox} focus on detecting text in scene images. Our proposed method to detect text in high entropy images is primarily inspired by both {\em Textbox++} and {\em Textbox}. {\em TextBox} and {\em TextBox++}~\cite{TB2016, TBPP2018} exhibit reasonably accurate performance in detecting text from scene images. 
\par
However, none of the existing techniques performs with high accuracy when it comes to detecting text from images where multiple texts are enclosed within regions along with possibly various other objects. This essentially means that the image has a high variability of heterogeneous information within compact regions. We relate this with the {\it entropy} of the image,  defined as the average information in an image and can be determined approximately from the histogram of multichannel and colour-space features of the image (see \cite{ENT1984}). In this work, we propose to detect text from such images by first segmenting it into multiple regions. The segmented regions are evaluated by first computing super-pixel level colour-space features and texture features from a dilated version of the image and hence, combing multiple super-pixels to form various segmented regions based on the super-pixel similarities. We propose an ensemble modelling approach for feature extraction by combining multiple Convolutional Neural Network (CNN) based models using selective non-max suppression. We use the ensemble model algorithm for detecting text of varying scales in each of the segments. 
\par
Our key contributions in this paper are two-fold: First, we designed a segmentation algorithm which is tailor-fitted to segment out the regions of the image containing text. Secondly, we propose an ensemble of multiple neural networks which extract features from segments of the image in order to detect text of varied sizes in the form of compact bounding boxes. The proposed architecture is highly parallelizable and the results show comparable and in cases, better accuracies in comparison to the competitors. 
\par
The rest of the paper is arranged as follows. In {\bf Section 1} we describe the framework of the image segmentation and the text detection in the form of a pipeline process. {\bf Section 2} is dedicated to describing the super-pixel based image segmentation methodology. The text detection method based on selective non-max suppression of the CNN ensemble is described in {\bf Section 3.} The implementation of the proposed framework along with the accuracy of the proposed method on various image datasets is discussed in {\bf Section 4.} We conclude the paper with a brief discussion in {\bf Section 5.}

\section{Framework}
\begin{figure}
\center

    \includegraphics[width=0.7\linewidth]{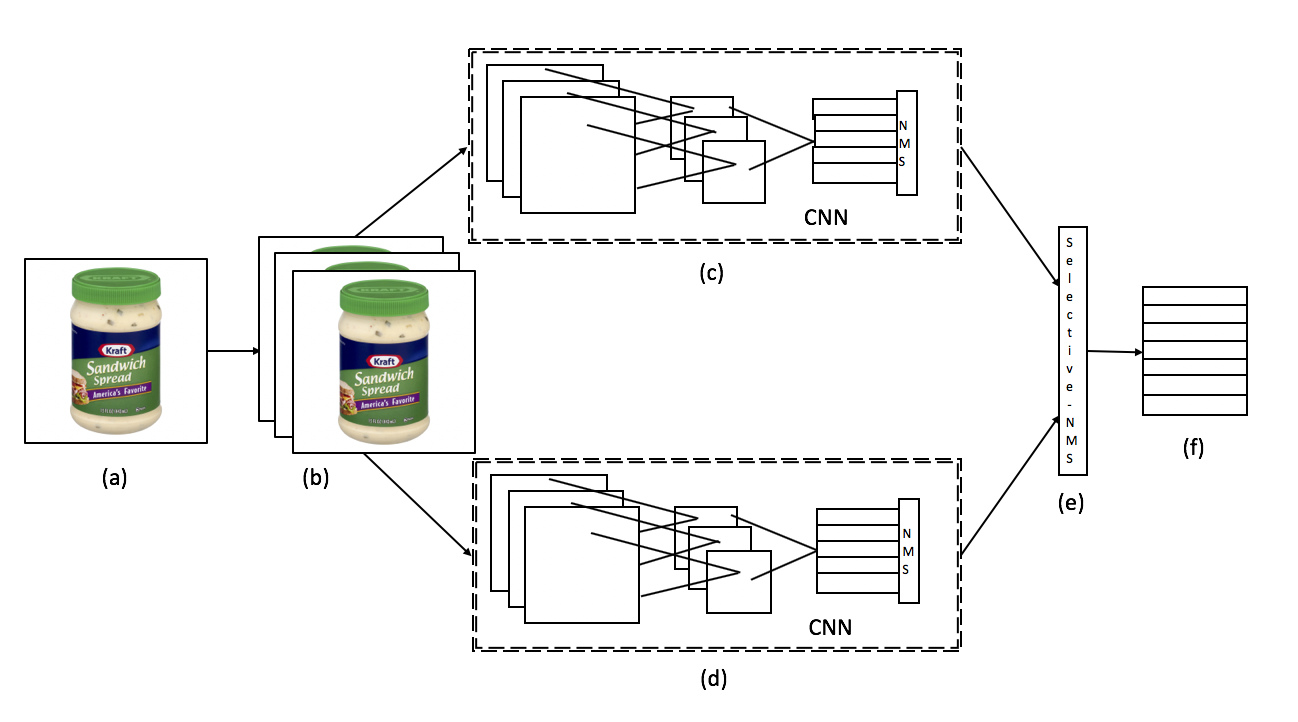}
    
    \caption{(a) Input image, (b) Segmented image, (c) and (d) TextBoxes\cite{TB2016} and TextBoxes++\cite{TBPP2018} CNN architectures trained on ICDAR2013, (e) Selective non-maximal suppression, (f) Final output text boxes.}
    \label{fig:arch}
\end{figure}

    The framework consists of two main components - (i) Image segmentation and  (ii)  Text detection. Image segmentation component splits the high entropy regions of the image into multiple segments so that it is easier for the subsequent text detection module to extract features from them more efficiently. The text detection component embodies a semi-bagging styled model. It is a pipeline process where the output of the image segmentation component is fed to the text detection component. The above components are elaborated more in the following sections. Fig.\ref{fig:arch} details the architecture of the complete framework end-to-end.

\section{Segmentation}
The proposed segmentation method consists of a number of intermediate steps resulting in spatially connected segments of homogeneous pixels. The goal of the segmentation module primarily is to make sure that the complete image is segmented into a number of regions such that the different text objects are enclosed in the individual regions. To ensure spatial continuity in the segments,
we first perform dilation of the image to make sure the holes and gaps inside objects are nullified and small intrusions at object boundaries are somewhat smoothened. We consider super-pixels of fixed size in the dilated image to calculate various features summarising super-pixel level information. Based on the super-pixel level feature information on the dilated image, we fit a Gaussian Mixture model to identify the class of super-pixels in an unsupervised manner. The details of the steps are described below.
\subsection{Dilation}
The image dilation is performed by convolving the image with a suitable kernel, in our case, the Gaussian Kernel. The anchor point of the kernel is chosen to be the centre of the kernel.  As we scan the chosen kernel over the image, the pixel value at the anchor point is replaced by the maximum pixel value of the image region overlapping the kernel. This results in the interesting regions of the image to grow and the holes and gaps within the object to get nullified. As a result, the segments having the objects are over-compensated making sure there is less chance of a segment to truncate objects inside its true boundaries and split at holes and gaps within the object.    
\subsection{Super-Pixel Features}
For each super-pixel of fixed size $s\in \mathcal S$, we calculate a set of features as $\bf x_s.$ For each super-pixel $s$ and each of the colour channels $c\in \mathcal C$, we calculate mean, standard deviation and energy, denoted by $\bf x^{(1)}_{s,c}=[\mu_{s,c}, \sigma_{s,c}, e_{s,c}].$ To summarize the texture features, we consider the Leung-Malik filterbank (see \cite{Leung})  at multiple scales and orientations. In total, we consider, first and second derivatives at 6 orientations, 8 Laplacians of Gaussian filters and 4 Gaussians and hence take the convolution with the pixel at different channels. To make sure there is orientation invariance, we take the maximum response over all orientations at each pixel. We calculate the mean, standard deviation and energy for all pixels within a super-pixel for all the filter convolution to get features 
$\bf x^{(2)}_{s,c}=[\mu_{s,c, j}, \sigma_{s,c, j}, e_{s,c, j}]_{j\in \mathcal J},$ for all colour channels $c\in \mathcal C$ and super-pixel $s\in \mathcal S$. The combined feature set for a given super-pixel is given by $\bf x_s=[\bf x^{(1)}_{s,c},x^{(2)}_{s,c}]_{c\in \mathcal C }.$
\begin{figure}
\begin{center}
    \includegraphics[width=0.6\linewidth]{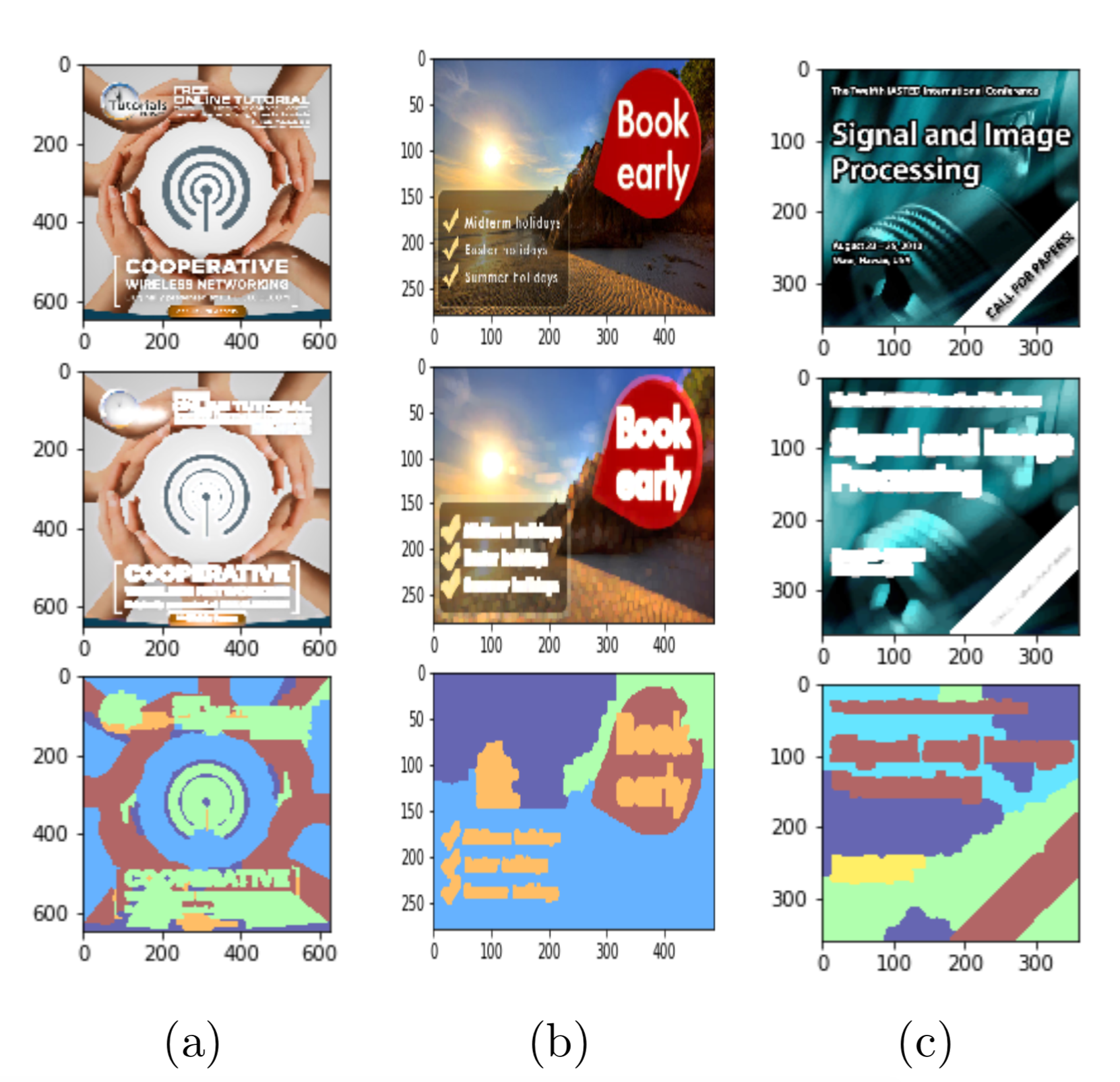}
\end{center}
\caption{The results of the proposed strategy of segmentation are shown for three different images in the three columns (a), (b) and (c). The actual image, the dilated image and the segmented image is shown for each of the examples in row 1, row 2 and row 3 respectively. }
\label{fig:seg}
\end{figure}

\subsection{Super-Pixel Similarity}
Following the strategy adopted by \cite{main}, we incorporate the similarity of neighbourhood super-pixels based on a function $w(s,s')$, for all $s,s'\in\mathcal S$. We combine information available over the entire set of features and spatial distance to calculate the similarity function w(.,.) between neighbourhood super-pixels. Denote the Euclidean distance between features of two super-pixels $s,s'$ by $d(\bf x_s, \bf x_{s'})$ and the standard deviation across all super-pixels by $\sigma_x$. The spatial Euclidean distance between a pair of super-pixels is given by $d(s, s')$ and the average distance across all super-pixels by $\overline d(\mathcal S).$ Combining the feature level information and spatial distance between super-pixels, the similarity function is given by $w(s, s')=exp(-\frac{d(\bf x_s, \bf x_{s'})}{2\sigma_x^2})\big(\frac{d(s,s')}{\overline d(\mathcal S)}\big)^{-1}.$

\subsection{Segment Classification}
Based on the computed features and weight function we classify the super-pixels into a number of classes in an unsupervised manner. Let us denote the unknown classes of the super-pixels by $Y=\{y_s, s\in \mathcal S\}$. If there are $K$ segments present in a given image, denoting $K$ classes, we have $y_s\in \{1,2,\ldots, K\}$ for $s\in \mathcal S.$ The class information given by the joint class probability function is factorized as $p(Y)=\prod_{s\in \mathcal S}\pi(y_s)\prod_{s,s'\in \mathcal S}R(y_s,y_s')$ where the class prior probabilities are given by $\pi(y_s).$ The  mutual information between a pair of neighbourhood super-pixels are given by $R(y_s,y_{s'})=\beta w(s,s')B(y_s,y_{s'}),$  $\beta>0$ being a tuning parameter controlling the spatial regularization. Here $B(y_s,y_{s'})$ is a spatial regularisation function indicating the chance of two neighbouring super-pixels to belong to the same class. We chose a diagonal structure of the matrix $[B(y_s,y_{s'}), s, s'\in\mathcal S]$ making all the diagonal elements to be identical to $1$. 

Given a fixed class $k$, the features are assumed to have a Gaussian distribution with fixed mean $\mu_k$ and variance-covariance matrix $\Sigma_k$ given by $p(\bf {x}_s|y_s=k)=N_k(\mu_k, \Sigma_k).$ Hence the super-pixel class is predicted by estimating the model parameters using the Expectation-Maximization algorithm and hence evaluating 
\begin{equation}
\begin{split}
(\hat{y}_s, s\in \mathcal S)=ArgMax_{{y}_s, s\in \mathcal S} \prod_{s\in \mathcal S}p(\bf {x}_s|y_s)\pi(y_s)\\\prod_{s,s'\in \mathcal S}R(y_s,y_{s'}).
\end{split}
\end{equation}
The estimated class information of the super-pixels is used to merge super-pixels of the same class level to get different segments. For three selected examples of images, the results of the proposed strategy after the dilation and segmentation is shown in Fig.\ref{fig:seg}. In the next section, we describe the text detection strategy where the ensemble of the CNN models is used to probe in each of the detected segments to extract texts of various sizes.

\section{Text Detection}
    The task of text detection in an image is very similar to object detection, where the text can be treated as an object. Hence, all object detection models can be used by making them binary classifiers - text (word level) and non-text. But all these object classifiers have their own limitations.
    \begin{itemize}
        \item Sometimes the image has a large amount of text compacted in a region forming a text cluster. Detecting these words separately becomes hard for conventional object detection techniques as they are trained to recognize a few numbers of separable objects in an image.
        \item Text in a single image can vary in both font-sizes and font-styles in a single image. Though it is claimed that most object detection methods are scale invariant, the results say otherwise as shown in Singh \etal ~\cite{Singh18}.
    \end{itemize}
    As shown in Liao \etal~\cite{TB2016, TBPP2018} having wide kernels ($3 \times 5$) rather than the generic square kernels help in better coverage in detecting text. Text in most cases, unlike objects, has a rectangular aspect ratio. Wide kernels will capture information about the aspect ratio of text objects better than square kernels. An ensemble of multiple CNN based models ensures a different level of information will be captured by different kind of models resulting in better coverage in information gathered from image.
    
    The models are then stitched together using selective non-max suppression algorithm. Non-Maximal Suppression removes multiple overlapping boxes detected for the same text location and keeps the one with the highest probability. Selective non-maximal suppression does the same but also takes into account the accuracy of the model from which the bounding box has been generated, giving it higher preference. Predictions from models which have a higher accuracy are preferred over others even if the individual probability might be slightly smaller.

    \subsection{Non-Maximal Suppression (NMS)}
        Let us assume that there are $n$ models and the number of bounding boxes predicted by $j^{th}$ model be $n_j$. Let $K$ be the list of all bounding boxes such that $k_{ij}$ is the $i^{th}$ bounding box predicted by model $j$ with $p_{k_{ij}}$ being the probability of that bounding box containing text. Let $\ell$ represent a sorted ordering of all these bounding boxes. That implies $\lvert \ell \rvert = \lvert K \rvert =  n_1 + n_2 + n_3 + ... + n_n$.

        \begin{algorithm}[H]
        \caption{NMS Algorithm}
        \label{NMSalgorithm}
        \begin{algorithmic}
        \Procedure{NMSalgorithm}{$k$, $p$, $nmsThreshold$}
            \State $\ell$ = sort($k$, $p$, $desc$) \Comment{sort $k$ based on prob. $p$ in desc order}
            \For{ $i = 1, ..., \lvert \ell \rvert $ }
                \For{ $ j = i+1, ... \lvert \ell \rvert $}
                    \If {$IOU(\ell_i, \ell_j) > nmsThreshold$}
                        \State $\ell.pop(j)$
                    \EndIf
                \EndFor
            \EndFor
            \State \Return {$\ell$}
    
        \EndProcedure
        \end{algorithmic}
        \end{algorithm}
    
    \subsection{Selective Non-Maximal Suppression (sNMS)}
        Let $M_q$ denote the model with the highest accuracy $a_q$ among all other models $M_i$ where $i \in \{1, 2, ..., n$\}. Let $P_t$ be the threshold probability - the probability that the bounding box is considered a true text box predicted by a model. $P_t$ is kept high for $M_q$, say $P_{t_h}$ while $P_t$ is kept slightly lower for the other $n-1$ models, say $P_{t_l}$. The bounding boxes predicted by each of the models are first filtered using this. After that, the probability of all the $n_q$ predicted boxes of model $M_q$ is assigned to $1$, while the probability of other boxes is left untouched. Post this reassignment of probabilities, NMS is performed on all the predicted boxes from all $n$ models.
    
        \begin{algorithm}[H]
        \caption{Selective NMS Algorithm}
        \label{SelectiveNMSalgorithm}
        \begin{algorithmic}
        \Procedure{SelectiveNMSalgorithm}{$a$, $P_{t_1}$, $P_{t_2}$, $k$, $p$, $nmsThreshold$}
            \State $q = max(a)$
            \For{ $k_q$ } \Comment{bounding boxes in $q^{th} model$}
                \State remove boxes where $p_q < P_{t_h}$
            \EndFor
            \For{ $k_q$ } \Comment{remaining bounding boxes in $q^{th} model$}
                \State $p_q = 1$ 
            \EndFor
            \For{ $ r = 1, ..., q-1, q+1, ..., m$}
                \For{ $k_r$}
                    \State remove boxes where $p_r < P_{t_l}$
                \EndFor
            \EndFor
            \State $bbs = $NMSalgorithm($k$, $p$, $nmsThreshold$)
            \State \Return {$bbs$}
    
        \EndProcedure
        \end{algorithmic}
        \end{algorithm}
    \vspace{-.2in}
        \paragraph{Notes and Comments.} Selective NMS ensures that the text boxes predicted with high probability by a model with the highest accuracy will always have priority over similar text boxes predicted with high probability by other models. 

\section{Implementation and Experiments}
    Our proposed methodology employs multiple models, to detect text from an image, which are stitched together using the selective-NMS algorithm. Multiple pre-trained models, as developed in Liao \etal~\cite{TB2016, TBPP2018} were used to detect text boxes from images. For selective-NMS, corresponding to the model with the highest accuracy of $0.9$ was set as a probability threshold above which a bounding box was considered a true text box with high confidence. The same parameter was set to 0.8 for the other models. NMS Threshold, the ratio of intersection area of two text boxes to union area of them (IOU), was set to $95\%$ i.e. with IOU above $95\%$ between two text boxes, they are considered to contain the same text. For text recognition, we synthesised 9 million text images using \textbf{SynthText} for various size, style and background of text for training. The full training set was run in a computer with standard K80 GPU and average execution time for detecting text in a single image is recorded to be around $0.15s$.
    \par
%-------------------------------------------------------------------------

%-------------------------------------------------------------------------

The ICDAR2013 dataset consists of images where the user is explicitly directing the focus of the camera on the text content of interest in a real scene. Product image dataset, on the other hand, consists of images of items taken from a high-resolution camera and have no background (white). By converting the image to grey-scale we calculate the entropy of the images in areas where the text is present. The average entropy of a sample of images from ICDAR2013 \cite{ICDAR2013} dataset was around $7.0$ while that of images from Walmart dataset \cite{WALMART2019} was around $6.0$ with $6.5$ marking a demarcation boundary for separating the two datasets. 

\subsection{Experimental Results}
\begin{figure}
\center
\includegraphics[width=0.8\linewidth]{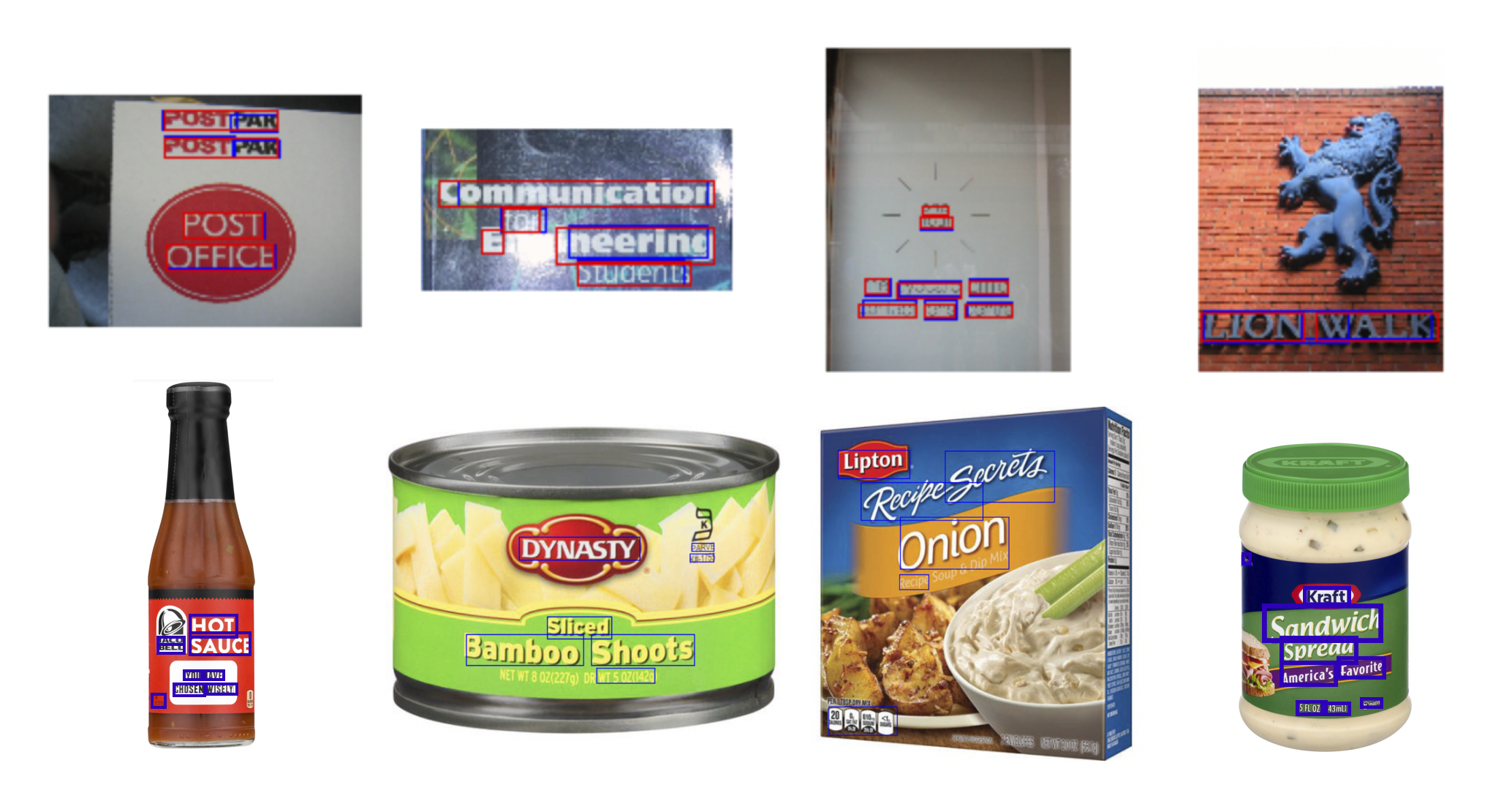}

\caption{Sample of results for text localization on ICDAR2013 dataset (top row) and high entropy image dataset (bottom row). Blue boxes represent the predicted bounding box while the red represent the ground truth bounding box.}
\label{fig:icdres}
\end{figure}

%ICDAR2013 dataset explanation
ICDAR2013 \cite{ICDAR2013} contains high-resolution real-world images. The models have been trained on the ICDAR2013 train set and then tested on the ICDAR2013 validation set. The results from all the models are then passed through selective NMS and the final bounding boxes are used for computing the metrics for precision, recall and f-score. Table.\ref{tab:comp} summarizes and compares our results with other methods. A sample of the results are shown in Fig.\ref{fig:icdres}. The predicted text regions are placed on top of the true text regions, indicated by blue and red boxes respectively.

\begin{table}[ht]
\centering
\begin{tabular}{|c|l|l|l|l|}
\hline
%\textbf{Datasets}  & \multicolumn{3}{c|}{\textbf{ICDAR2013}} &  \multirow{2}{*}{\textbf{Time/s}} 
%\cline{1-2}\cline{2-4} 
%{\textbf{Methods}} & \textbf{P}  &\textbf{R}   &\textbf{F}    &       \\ \hline                                                                 
\textbf{Datasets}  &  \multicolumn{3}{c|}{\textbf{ICDAR2013}} & \textbf{Time/s}  \\ \hline
\textbf{Methods} & \textbf{P}  &\textbf{R}   &\textbf{F} & \\ \hline

\shortstack{MMser\cite{mmser}\\(Zamberletti, Noce, and Gallo 2014)}                  &0.86         &0.70         &0.77                   &0.75    \\ \hline
\shortstack{TextFlow\cite{textflow}\\ (Tian et al. 2015)}                              &0.85         &0.76         &0.80                   &1.4    \\ \hline
\shortstack{FCN\cite{FCN}\\ (Zhang et al. 2016)}                                  &0.88         &0.78         &0.83                   &2.1    \\ \hline
\shortstack{SSD\cite{ssd}\\ (Liu et al. 2016)}                                    &0.80         &0.60         &0.68                   &0.1    \\ \hline
TextBoxes\cite{TB2016}                                                                &0.86         &0.74         &0.80                   &0.09    \\ \hline
TextBoxes++\cite{TBPP2018}                                                              &0.86         &0.74         &0.80                   &0.10    \\ \hline
Ensemble                                                                 &\textbf{0.83}         &\textbf{0.77}         &\textbf{0.80}                   &\textbf{0.15}    \\ \hline

\end{tabular}
\caption{Text localization on ICDAR2013. P, R and F refer to precision, recall and F-measure respectively.}
\label{tab:comp}
%\vspace{-1em}
\end{table}

\par
%Walmart Dataset explanation
The ensemble model is also tested on a dataset containing publicly available product images on the Walmart website \cite{WALMART2019}.  These are high resolution and high entropy images of the front face of processed food items used on a daily basis by consumers. The predicted text region bounding boxes enclose regions containing texts of multiple sizes and mixed font types in the same image. This is particularly important for product images as the product labels often contain texts of multiple fonts. The proposed text detection strategy also successfully detects text regions when the text is moderately rotated or curved due to the shape of the product package, eg- a can or a bottle (see the 2nd, 3rd and 4th images in the bottom row in Fig.\ref{fig:icdres} ). The use of wide kernels turn out to be useful in detecting horizontal text boxes and on top of it, the image segmentation and CNN ensemble network consider image convolution filters at multiple scales and rotation angles. This contributes to ensuring that the text box detection accuracy is invariant at least under limited distortion and rotation of the horizontal orientation of the texts. 
\par
The models trained on ICDAR2013 train set are used on 50 images from this dataset where the ground truth boxes are known. The main difference between the images in this dataset and the other publicly available datasets is that the images have no background noise that is usually present in scene text. However multiple texts are usually present in a small region of the image along with various other objects resulting in high local entropy. Most of the models currently available perform poorly on detecting text in such regions in the image. In such cases, the model described in this paper performs better than the existing ones in terms of precision, recall as well as f-score. Fig.\ref{fig:icdres} gives the results achieved on the datasets ~\cite{WALMART2019, ICDAR2013} by the model. In the case of ICDAR2013 dataset, the model has performed at par with the existing models currently available, but this improves drastically in the case of the dataset containing high entropy images ~\cite{WALMART2019}. The precision is at least $6\%$ higher than the existing methods while recall is higher by around $15\%$. Table.\ref{tab:wal} compares the results achieved on the Walmart High Entropy Images.
\begin{table}[ht]
\centering
\begin{tabular}{|c|l|l|l|}
\hline
\textbf{Datasets}   & \multicolumn{3}{c|}{\textbf{High Entropy Images ~\cite{WALMART2019}}} \\ \hline
\textbf{Methods}   & \textbf{P}  &\textbf{R}   &\textbf{F}     \\ \hline
TextBoxes                                                                &0.867         &0.264         &0.405     \\ \hline
TextBoxes++                                                                &0.831         &0.311         &0.453     \\ \hline
Ensemble                                                                 &\textbf{0.920}         &\textbf{0.467}         &\textbf{0.619}     \\ \hline

\end{tabular}
\caption{Text localization results on High entropy image dataset. P, R and F refer to precision, recall and F-measure respectively.}
\label{tab:wal}
%\vspace{-1em}
\end{table}
\vspace{-0.2in}
%Figure 1 - ICDAR test results
%Figure 2 - Walmart Test results
%Figure 3, 4 - Segmentation images - Ensemble result and seg result 
%------------------------------------------------------------------------
\section{Conclusion}
We have presented an algorithm which employs an ensemble of multiple fully convolutional networks preceded by an image segmenter for text detection. It is highly stable and parallelizable. It can detect words of varied sizes in an image which is very high on entropy. Comprehensive evaluations and comparisons on benchmark datasets clearly validate the advantages of our method in three related tasks including text detection, word spotting and end-to-end recognition. It even exhibits better performance than {\em TextBox} and {\em TextBox++}~\cite{TB2016, TBPP2018} in detecting graphical text in an image. The ICDAR2013 dataset images have real-world contents and background noise surrounding the true text regions, unlike the Walmart High Entropy Images, where the challenge is largely the presence of multiple textual elements within small regions resulting higher entropy. The proposed methodology is particularly targeted to work on such high entropy text regions and hence performs very well on Walmart High Entropy Images. However, a more targeted background removal strategy, image segmentation and text candidate pre-filtering using text region specific key point identification and feature descriptions such as Stroke width descriptors, Maximally Stable Extremal Region descriptors should enhance the performance of the CNN ensemble model more. We keep this development for future communication. 

\bibliography{bmvc_final}

\end{document}